\documentclass[3p]{elsarticle}

\usepackage{lineno}
\usepackage[hidelinks]{hyperref}

\modulolinenumbers[5]

\usepackage{amsfonts}
\usepackage{xcolor} 
\usepackage[utf8]{inputenc}
\usepackage{csquotes}
\usepackage{graphicx} 
\usepackage{wrapfig}
\usepackage{listings}
\usepackage{float}
\usepackage{color}
\usepackage{xcolor} 

\definecolor{dkgreen}{rgb}{0,0.6,0}
\definecolor{gray}{rgb}{0.5,0.5,0.5}
\definecolor{mauve}{rgb}{0.58,0,0.82}
\definecolor{mygreen}{rgb}{0,0.6,0}
\definecolor{mygray}{rgb}{0.5,0.5,0.5}
\definecolor{mymauve}{rgb}{0.58,0,0.82}
\definecolor{codegreen}{rgb}{0,0.4,0}
\definecolor{codegray}{rgb}{0.5,0.5,0.5}
\definecolor{codepurple}{rgb}{0.58,0,0.82}
\definecolor{cyan}{rgb}{0,0.8,0.8}
\definecolor{steelblue}{RGB}{70,130,180}
\definecolor{gray60}{rgb}{0.6, 0.6, 0.6}
\definecolor{backcolour}{rgb}{0.94,0.94,0.94}
\definecolor{purpleU}{RGB}{48,10,36}
\definecolor{mPurple}{rgb}{0.58,0,0.82}

\lstdefinestyle{Python}{ 
	language=Python,
    backgroundcolor=\color{white},
    commentstyle=\color{gray},
    keywordstyle=\bfseries\color{blue},
    numberstyle=\tiny\color{mygray},
    stringstyle=\color{red},
    basicstyle=\footnotesize,
    breakatwhitespace=false,
    breaklines=true,
    keepspaces=true,
    numbers=left,
    numbersep=5pt,
    showspaces=false,
    showstringspaces=false,
    showtabs=false,
    tabsize=2,
    basicstyle={\small\ttfamily},
}
\makeatletter
\def\ps@pprintTitle{%
 \let\@oddhead\@empty
 \let\@evenhead\@empty
 \def\@oddfoot{}%
 \let\@evenfoot\@oddfoot}
\makeatother

\begin{document}
\begin{frontmatter}
\title{MOGPTK: The Multi-Output Gaussian Process Toolkit}
\author{Taco de Wolff, Alejandro Cuevas, Felipe Tobar \vspace{0.5em}\\ \texttt{\{tdewolff,acuevas,ftobar\}@dim.uchile.cl}}

\address{Center for Mathematical Modeling\\
       Universidad de Chile}

\begin{abstract}
We present MOGPTK, a Python package for multi-channel data modelling using Gaussian processes (GP). The aim of this toolkit is to make multi-output GP (MOGP) models accessible to researchers, data scientists, and practitioners alike. MOGPTK uses a Python front-end, relies on the GPflow suite and is built on a TensorFlow back-end, thus enabling GPU-accelerated training. The toolkit facilitates implementing the entire pipeline of GP modelling, including data loading, parameter initialization, model learning, parameter interpretation, up to data imputation and extrapolation. MOGPTK implements the main multi-output covariance kernels from literature, as well as spectral-based parameter initialization strategies. The source code, tutorials and examples in the form of Jupyter notebooks, together with the API documentation, can be found at \url{http://github.com/GAMES-UChile/mogptk}.
\end{abstract}


\end{frontmatter}

\section{Introduction}
\label{sec:intro}
The Gaussian process (GP) is a Bayesian nonparametric model for time series, that has had a significant impact in the machine learning community following the seminal publication of \citep{rasmussen_gaussian_2006}. GPs are designed through parametrizing a covariance kernel, meaning that constructing expressive kernels allows for an improved representation of complex signals. 
Recent advances extend the GP concept to multiple series (or channels), where both auto-correlations and cross-correlations among channels are designed jointly; we refer to these models as multi-output GP (MOGP) models. A key attribute of MOGPs is that appropriate cross-correlations allow for improved data-imputation and prediction tasks when the channels have missing data. Popular MOGP models include: i) the Linear Model of Coregionalization (LMC) \citep{goovaerts_1997}, ii) the Cross-Spectral Mixture (CSM) \citep{ulrich2016gaussian}, iii) the Convolutional Model (CONV) \citep{alvarez_2009}, and iv) the Multi-Output Spectral Mixture (MOSM) \citep{parra_spectral_2017}. Training MOGPs is challenging due to the large number of parameters required to model all the cross-correlations, and the fact that most of MOGP models are parametrized in the spectral domain, thus being prone to local minima. Therefore, a unified framework that implements these MOGPs is required both by the the GP research community as well as by those interested in practical applications for multi-channel data.

The multi-output Gaussian process toolkit (MOGPTK) aims to address the need for an MOGP computational toolkit in the form of a Python package that implements the mentioned MOGP kernels and provides a natural way to train and use them. MOGPTK is built upon GPflow~\citep{GPflow2017}, an extensive GP framework with a wide variety of implemented kernels, likelihoods and training strategies. GPflow is in turn built upon TensorFlow~\citep{abadi2016tensorflow}, a framework that allows for the construction of computational graphs of tensors and operations which can be calculated on either CPUs or GPUs. Needless to say, GPU-training is much desired due to the ability of graphics cards to perform linear operations in parallel.

\section{Existing MOGP libraries and scope of MOGPTK}
Previous toolkits for MOGPs include GPmat~\citep{GPmat} (University of Sheffield) through its module called \texttt{multigp}, a MATLAB library that includes sparse approximations and implements multi-output support through convolution processes~\citep{alvarez_2009}. Another library is GPy~\citep{gpy2014} (University of Sheffield), a Python package that implements the Intrinsic Model of Coregionalization (IMC) and LMC kernels. More recently, GPyTorch (Cornell University) is a Python library for general GP modelling that uses PyTorch to facilitate faster training on GPUs~\citep{gpytorch}. GPyTorch implements the LMC kernel and the multi-task kernel by~\cite{bonilla2008}. Lastly, GPflow, the framework upon which our work is based, also has multi-output support using the LMC kernel~\citep{GPflow2017}.

Neither of the above libraries implement the---by now standard---CSM, CONV or MOSM models described in Section \ref{sec:intro}. Critically, not all libraries even allow for different numbers of data points per channel. Furthermore, existing libraries give little emphasis on improving training through parameter initialization and they usually lack parameter interpretation. MOGPTK, conversely, facilitates the whole process of implementing an MOGP, from data loading and parameter initialization to model training and interpretation. Our toolkit also implements all main MOGP models mentioned in Section \ref{sec:intro}.

\section{Functionality}
The main pillars of MOGPTK are the included MOGP models, data handling, parameter initialization and parameter interpretation, each discussed below.

\subsection{Models}
MOGPTK considers a base MOGP kernel from which specific kernels are derived. The base kernel provides the functionality to split the input data into multiple channels and process them by sub-kernels. While single-channel kernels implemented in GPflow have input data of shape $N\times P$ (with $N\in \mathbb{N}$ the total number of data points and $P\in \mathbb{N}$ the number of input dimensions), the MOGPTK base kernel has input data of shape $N\times (1+P)$, where the first column contains integers denoting the channel index to which the remaining columns correspond. Then, using the channel indices the base kernel splits the data into its different channels for the sub-kernels to operate. This allows us to manage different amounts of data points per channel, therefore, the total amount of data points can be express as $N = \sum_{m=0}^M N_m$, with $M$ the number of channels and $N_m$ the number of data points in channel $m$.



\subsection{Data handling}
MOGPTK features general-purpose classes to perform common data-analysis operations effortlessly. Data can be loaded from various sources (e.g., CSV files, Pandas DataFrames, or generated using Python functions) and formatted if necessary. For instance, data containing date and/or time values can be automatically converted to structured (numerical) representations for compatibility with the rest of the toolkit. Data can also be pre-processed using included transformations such as detrending or logarithm among others, which can be applied in compositional manner in order for the models to be trained effectively. After training, the transformation can be reverted to the original domain in the same vein of \cite{rios_tobar2019}. Additionally, MOGPTK allows for removing data ranges to simulate missing data or sensor failure and the data can be easily plotted in time or spectral domain.

\subsection{Parameter initialization}

Training MOGPs can be challenging due to their large number of hyperparameters and highly complex objective function. In this sense, MOGPTK features two methods for setting appropriate initial conditions for the  hyperparameters. The first one is based on training independent spectral mixture kernels \citep{wilson_gaussian_2013} to individual channels and using their spectral Gaussian means and variances to compute the initial parameter values for the multi-output kernels. The second method utilizes the Bayesian Nonparametric Spectral Estimation (BNSE)~\citep{tobar2018bayesian}, or the Lomb-Scargle method, to identify hyperparametes from the channels spectral content. Both methods are single-output and thus are trained independently on each channel, therefore, subsequent (maximum likelihood) model training can focus on training the inter-channel cross-correlations.

\subsection{Parameter interpretation}


Besides training and prediction, MOGPTK also provides interpretation of hyperparameter values via visualization techniques. MOGPTK shows the correlations between channels for different kernels, in the particular case of spectral kernels (e.g., SM, MOSM, CSM, SM-LMC), this reveals the cross-spectral coupling between channels. For instance, when comparing stock values in financial applications we can usually observe correlating quarterly patterns due to the anticipated quarterly reports by stock investors as shown in \cite{dewolff_cuevas_tobar2020}. Additionally, retail markets may also correlate monthly due to salaries being paid at that frequency and thus producing a spike in sales. The interpretation of such relationships is relevant for practical applications that require deeper understanding and analysis beyond mere imputation and extrapolation of data.

\section{Example}
We next provide a short example of how MOGPTK operates on an air-quality time series of four channels. The dataset \cite{DEVITO2008750} contains hourly-average measurements of five metal oxide chemical sensors embedded in an \emph{Air Quality Chemical Multisensor}. The data were collected in a polluted Italian city between March 2004 and February 2005. We considered the MOSM kernel, with three spectral components per channel, initialized using BNSE and optimized using BFGS.

\begin{figure}[H]
    \centering
    \includegraphics[width=0.8\linewidth]{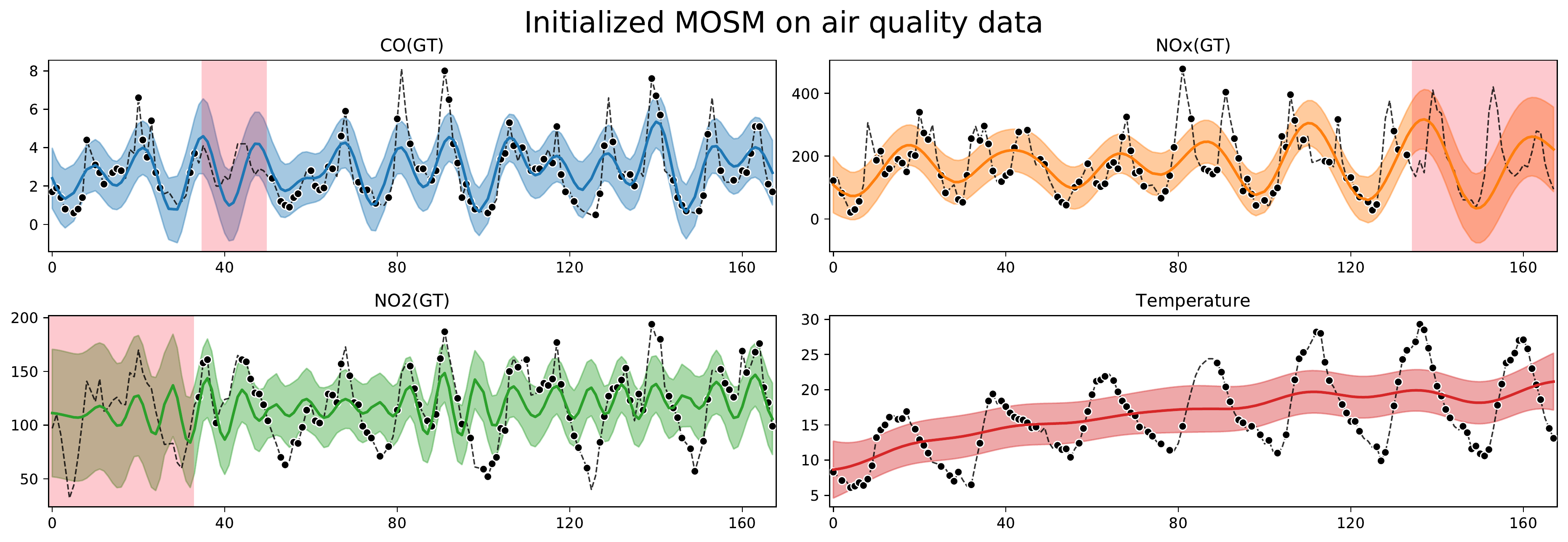}
    \caption{MOSM kernel prediction using parameters initialized by BNSE, i.e., no likelihood optimization has been performed at this stage. Notice how (uncorrelated) fundamental frequency components can be identified in each channel.}
    \label{fig:initialization}
\end{figure}

\begin{figure}[H]
    \centering
    \includegraphics[width=0.8\linewidth]{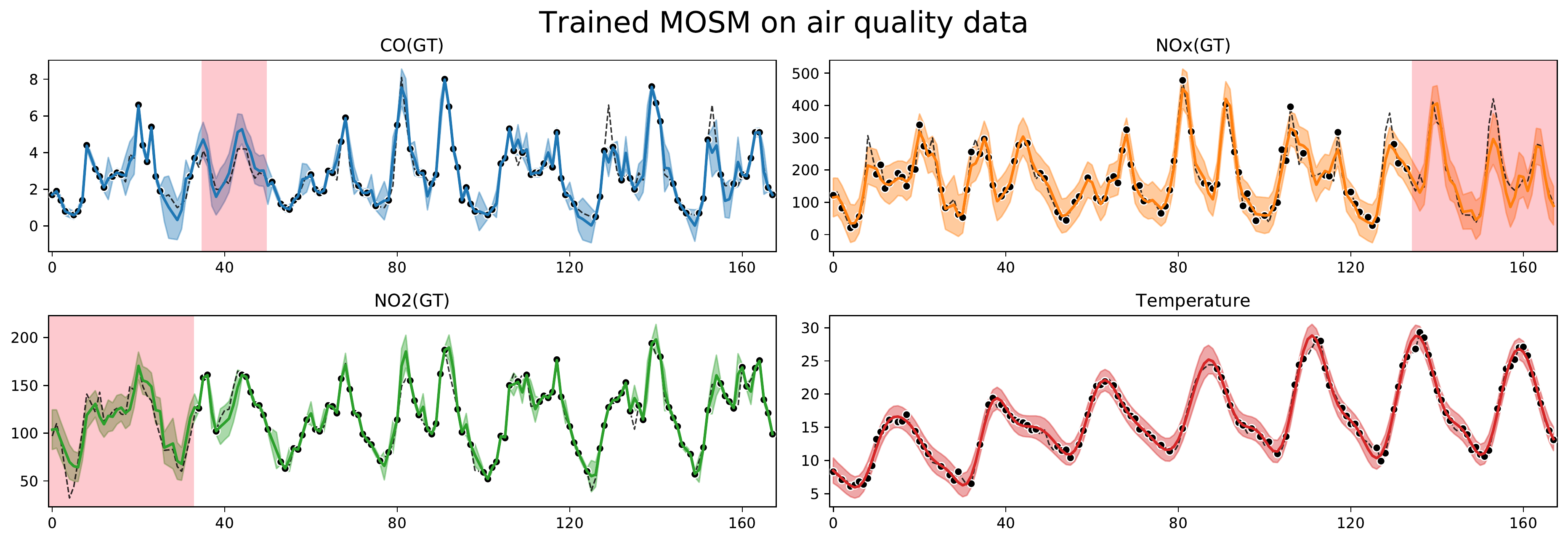}
    \caption{MOSM kernel prediction. Using BNSE parameter initialization as initial condition, the MOSM kernel was trained via maximum likelihood using L-BFGS-B with 500 iterations.}
    \label{fig:training}
\end{figure}

Listing~\ref{lst:example} shows the corresponding code. We first loaded the (pre-processed) dataset into MOGPTK creating a four-channel model, then for each channel we removed a range of data to simulate sensor failure and additionally removed 30\% of the data points randomly. For improved training results, the internal data representation for all channels were linearly detrended and normalized to have zero mean and unit variance using the transformations provided in the toolkit. Next, we set up the MOSM kernel and initialized the parameters using BNSE. Fig.~\ref{fig:initialization} shows the results of solely initialization (not training), where only the main sinusoidal components can be identified. Lastly, the MOSM model was trained using the L-BFGS-B optimizer. With its three components per channel, MOSM featured 65 hyperparameter to train using 439 training points. Training took less than two minutes on an average CPU and was faster when utilizing a GPU. Fig.~\ref{fig:training} shows the results of the trained MOSM kernel, where the model was able to accurately interpolate and extrapolate beyond the observation range.

\begin{center}
\begin{minipage}[t]{0.8\linewidth}
\begin{lstlisting}[caption={Implementation of MOGPTK of a four-channel air-quality dataset: data loading, parameter initialization, model training and prediction.},captionpos=b,label={lst:example}]
import mogptk

# Load the pre-processed dataset into MOGPTK
x_col = 'Time'
y_col = ['CO(GT)', 'NMHC(GT)', 'NOx(GT)', 'NO2(GT)']
data = mogptk.LoadCSV('data/air_quality.csv', x_col=x_col, y_col=y_col)

# Remove ranges to simulate sensor failure
data[0].remove_relative_range(0.2, 0.3)
data[1].remove_relative_range(0.8, 1.0)
data[2].remove_relative_range(0.8, 1.0)
data[3].remove_relative_range(0.0, 0.2)

# Randomly remove points, detrend and whiten (mean=0,var=1)
for channel in data:
    channel.remove_randomly(pct=0.3)
    channel.transform(mogptk.TransformDetrend(degree=1))
    channel.transform(mogptk.TransformWhiten())

# Initialize parameters using BNSE
mosm = mogptk.MOSM(data, Q=3)
mosm.init_parameters('BNSE')
mosm.predict()
mosm.plot_prediction()

# Optimize parameters using L-BFGS-B
mosm.train('L-BFGS-B')
mosm.predict()
mosm.plot_prediction()
\end{lstlisting}
\end{minipage}
\end{center}

\section{Availability and documentation}
MOGPTK is released under the MIT license and thus it can be used in both open-source and commercial applications. The source code is publicly available on GitHub at \url{https://github.com/GAMES-UChile/mogptk/}, where contributions are encouraged and issues can be raised. The repository contains tutorials and examples with real-word data in the form of Jupyter notebooks. Additionally, the API documentation describing all methods can be accessed at \url{https://games-uchile.github.io/mogptk}. MOGPTK requires atleast Python 3.6 and TensorFlow 2, and can be installed executing \texttt{pip install mogptk}.

\section*{Acknowledgements}
We are thankful to the Center for Mathematical Modeling (Conicyt AFB \#170001), without its invaluable support MOGPTK would only be an idea. We also thank  Fondecyt-Iniciación \#11171165. We would like to thank Cristóbal Silva, Gabriel Parra, Mario Garrido, and Victor Caro for their feedback on earlier versions of MOGPTK. This document has been formatted using Elsevier's \emph{elsarticle.cls} \LaTeX~document class.

\bibliography{refs}
\end{document}